\documentclass[10pt, conference, compsocconf]{IEEEtran}
\newlength{\bibitemsep}
\newlength{\bibparskip}
\let\oldthebibliography\thebibliography
\renewcommand\thebibliography[1]{%
  \oldthebibliography{#1}%
  \setlength{\parskip}{\bibitemsep}%
  \setlength{\itemsep}{\bibparskip}%
}

\ifCLASSINFOpdf
\else
\fi

\usepackage{graphicx}
\usepackage{amsmath}
\usepackage{amssymb}
\usepackage{booktabs}
\usepackage{flushend}
\usepackage{comment}
\usepackage{multirow}
\usepackage[pagebackref,breaklinks,colorlinks]{hyperref}

\usepackage[capitalize]{cleveref}

\begin{document}
\title{Adaptive Memory Management for Video Object Segmentation}


\author{\IEEEauthorblockN{Ali Pourganjalikhan, Charalambos Poullis}
\IEEEauthorblockA{Immersive and Creative Technologies Lab,
Department of Computer Science and Software Engineering\\
Concordia University, Montreal, Canada\\
apourganjalikhan@gmail.com, charalambos@poullis.org}
% \and
% \IEEEauthorblockN{Charalambos Poullis}
% \IEEEauthorblockA{Gina Cody School of Engineering and Computer Science\\
% Concordia University\\
% Montreal, Canada\\
% Email: charalambos@poullis.org}
}


%Telephone: (800) 555--1212, Fax: (888) 555--1212}


\maketitle

\begin{abstract}
Matching-based networks have achieved state-of-the-art performance for video object segmentation (VOS) tasks by storing every-\textit{k} frames in an external memory bank for future inference. Storing the intermediate frames' predictions provides the network with richer cues for segmenting an object in the current frame. However, the size of the memory bank gradually increases with the length of the video, which slows down inference speed and makes it impractical to handle arbitrary length videos.

This paper proposes an adaptive memory bank strategy for matching-based networks for semi-supervised video object segmentation (VOS) that can handle videos of arbitrary length by discarding obsolete features. Features are indexed based on their importance in the segmentation of the objects in previous frames. Based on the index, we discard unimportant features to accommodate new features. We present our experiments on DAVIS 2016, DAVIS 2017, and Youtube-VOS that demonstrate that our method outperforms state-of-the-art that employ first-and-latest strategy with fixed-sized memory banks and achieves comparable performance to the every-\textit{k} strategy with increasing-sized memory banks. Furthermore, experiments show that our method increases inference speed by up to \textbf{80\%} over the every-\textit{k} and \textbf{35\%} over first-and-latest strategies.
\end{abstract}


% \fi
\IEEEpeerreviewmaketitle

\vspace{-10pt}
\section{Introduction}
\vspace{-5pt}
Video object segmentation (VOS) is a fundamental computer vision task with many applications in self-driving cars \cite{badue2021self}, augmented reality \cite{ngan2011video}, video editing \cite{bertolino2014sensarea}, and many other video-related tasks. Additionally, video object segmentation serves as a building block in tasks such as interactive video object segmentation \cite{cheng2021mivos,oh2019fast,miao2020memory,liang2020memory} and video instance segmentation \cite{feng2019dual,yang2019video}. 

In VOS, the target objects are annotated in their first appearance, and the objective is to segment them in subsequent frames. Early attempts on VOS \cite{caelles2017one,voigtlaender2017online,luiten2018premvos}, fine-tuned the network to learn the target objects' appearance. Fine-tuning a deep neural network with only one example (one-shot) at the target is challenging \cite{finn2017model,yang2018efficient}. Moreover, fine-tuning makes inference slow which makes it unsuitable for real-time applications. 

Recent works \cite{oh2019video,voigtlaender2019feelvos,yang2020collaborative}, instead of learning object features implicitly, use a learned embedding space to embed and memorize object appearance and use that embedding to segment object in the subsequent frames by calculating affinity between current and past frames embeddings. 

A key challenge in matching-based VOS is exploiting the previous frames' information. Using all previous frames' information is impracticable and redundant. Most recent works \cite{oh2018fast,voigtlaender2019feelvos,yang2020collaborative} rely only on the first-and-latest frame. The latest frame is visually close to the current frame, and the first frame provides reliable annotation for the object, avoiding drifting during the segmentation. This strategy disregards all intermediate frames' information. 

To address this, Space-Time Memory (STM) network \cite{oh2019video} employed a memory bank to store every-\textit{$\kappa$} frame for subsequent inferences. While this method utilizes intermediate frames, information of the memory bank grows linearly as a function of the number of frames $\kappa$. Hence, it imposes significant memory requirements that prevent the processing of longer video sequences. The solution is to remove features from the memory bank when they become obsolete, ensuring a fixed-sized memory bank that allows the processing of videos of arbitrary length. 

In this paper, we propose Least-Frequently-Used (LFU) feature removal based on the top-k index. Our method outperforms the first-and-latest strategy and achieves comparable results with the every-\textit{k} sampling strategy while using a smaller, fixed-sized memory bank. In addition to handling arbitrary-length video sequences without imposing significant memory requirements, our method considerably improves the inference speed when compared to the top two sampling strategies, first-and-latest and every-\textit{k}.

\begin{figure*}[ht]
    \centering
  \includegraphics[width=0.9\textwidth]{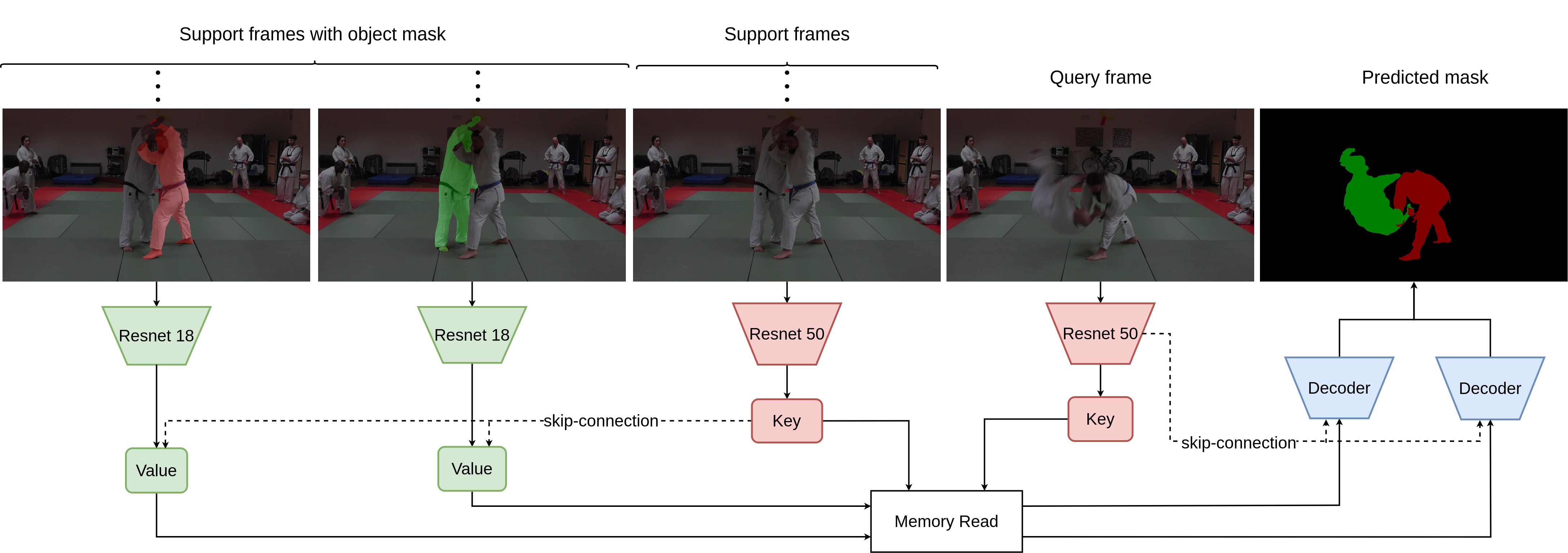}
  \caption{Overview of the network architecture. Parts with the same colour have shared weights. We illustrate the memory bank architecture in detail in Figure \ref{fig:memory}. Deep features learned by the key encoder are concatenated to the value encoder's input. The output at different levels of the key encoder is used to refine the mask at different scales of the decoder. We store the query key of specific frames for further inference. In the case of multiple objects, soft-aggregation is used to reach the final output.}
  \label{fig:overall}
\end{figure*}

\vspace{-5pt}
\section{Related Work}
\vspace{-5pt}
In this section, we review the most relevant works in video object segmentation.
\noindent
\textbf{Semi-supervised video object segmentation.} Following the taxonomy proposed by \cite{liang2020video}, recent VOS methods can be categorized into implicit and explicit according to the approach followed to address the problem. 

Implicit models aim at learning objects' representation by fine-tuning network weights for each object. Detection-based implicit methods such as \cite{maninis2018video, caelles2017one} segment objects in each frame independently without enforcing temporal consistency between consecutive frames. Propagation-based implicit methods such as \cite{perazzi2017learning, hu2018maskrnn, khoreva2017lucid, li2018video} propagate objects' masks between consecutive frames. 

Explicit models mainly rely on online learning to adapt to different objects. The online learning process makes these methods very time-consuming during the inference and reduces the network's ability to handle deformation. On the other hand, explicit models use a fixed set of parameters for all sequences during the inference. The features of the object from the previous frames are used to classify the current frame pixels as foreground or background using similarity matching; hence often called matching-based methods as well. A matching-based network uses different sources of information. \cite{oh2018fast,voigtlaender2019feelvos,wang2019ranet,lin2019agss} utilize information from first-and-latest frames to segment the object in the current image. \cite{oh2019video} uses a memory bank to store the object representation in intermediate frames. Despite the difference in architecture and sources of information, when it comes to incorporating previous frames' information into the current frame segmentation process, most of the approaches use cosine or Euclidean distances to find the affinity between previous frames and current frames features.

\noindent
\textbf{Memory Banks.} Networks can learn to read and write helpful information in external memory. For example, end-to-end memory networks have proven useful for document Q\&A \cite{kumar2016ask,miller2016key,sukhbaatar2015end}, visual tracking \cite{yang2018learning}, video understanding \cite{na2017read}, and summarization \cite{lee2018memory}. 

STM \cite{oh2019video} encodes each image into the key-value pair and uses them as cues to segment the target object. In their framework, past frames are in memory, and the current frame forms the query. Using the keys, they estimate the affinity between memory and query frame to determine which memory values should be used during segmentation.  Space-Time Correspondence Networks (STCN) \cite{cheng2021rethinking} uses object-agnostic keys, which reduce the computation in multi-object scenarios.

STM \cite{oh2019video} and its extensions and variants \cite{seong2020kernelized,li2020fast,hu2021learning,xie2021efficient,wang2021swiftnet,zhang2020spatial,zhou2019enhanced} store intermediate frames' key-value into a memory bank. The efficiency of the memory bank depends on the number of key-value pairs that can be stored. STM \cite{oh2019video} adds features for every-\textit{k} frame ($k=5$ in their paper) in the memory bank. 

Although increasing the number of frames in the memory bank improves performance,  MiVOS \cite{cheng2021mivos} showed that only a few of the memory features meaningfully contribute to the segmentation process. Instead, MiVOS uses only the k-closest memory features for each query feature and discards the rest as they adversely affect the performance. Although MiVOS \cite{cheng2021mivos} ignores redundant features during inference, it still maintains these features in the memory bank for future inferences.

\cite{liang2020video} propose an update and remove strategy to keep the memory bank size fixed. When adding new features to the memory bank, they perform an eligibility check for an update by ensuring that the feature's distance to its closest neighbour is lower than a threshold. The update is a running average between new features and their closest neighbour in the memory. To remove an obsolete feature that can not be merged, they keep track of its frequency being close to a query feature and then use a least-frequently-used strategy to remove it. Although \cite{liang2020video} keeps the size of the memory bank fixed, both update and removal steps use a threshold that is dependent on an affinity metric. The affinity metric for this approach needs to be bounded since the threshold needs are decided beforehand.

\vspace{-10pt}
\section{Methodology}
\vspace{-5pt}
Given a sequence of frames and the masks of the target objects in their first appearance (objects typically appear in the first frame; however, new objects can also appear in the middle of the sequence), we segment the object in the rest of the frame sequence. This problem is a variant of one-shot image segmentation \cite{shaban2017one} in which the current frame is the \textbf{query} and the past frames with segmented objects mask are the \textbf{support} set. Figure \ref{fig:overall} shows an overview of the proposed technique. We have chosen STCN \cite{cheng2021rethinking} as the baseline since it is effective and minimal. Although we evaluate our method on STCN, the feature sampling strategy can be used for any matching-based network with a memory bank. For each query and support frame, we extract a key that is independent of the object. The affinity between the query and support keys is then used to select the corresponding support values for segmentation.

\subsection{Encoding of key-value}
\vspace{-5pt}
Unlike STM \cite{oh2019video}, STCN\cite{cheng2021rethinking}, we use object agnostic key encoding with shared weights for support and query. We employ a Resnet50 \cite{he2016deep} followed by a $3 \times 3$ convolution layer as a projection layer to encode the key. Spatial information is preserved by using the output of \textit{res4} layer for the projection. The projection layer reduces the number of channels from $1024$ to $64$. The query key can be reused as a support key since the query and support key encoder have shared weights.

Value encoding is performed more frequently; therefore, we use a lighter backbone. We employ Resnet18 to encode the image and mask of an object. Unlike key encoding, value encoding is object-specific and only used for support frames. The output is then concatenated with the corresponding feature map from the key encoder and processed by two residual blocks. In this way, the network can use deeper backbone feature embeddings without any overhead. 

We initialize both key and value encoders with pre-trained weights from Imagenet \cite{deng2009imagenet}. In the value encoder, the input consists of an image and a corresponding mask. Hence, we modify the first layer of the Resnet18 to have 4-channels and initialize the additional new weights to zero.

\subsection{Memory read}
\begin{figure}[!ht]
    \centering
  \includegraphics[width=0.4\textwidth]{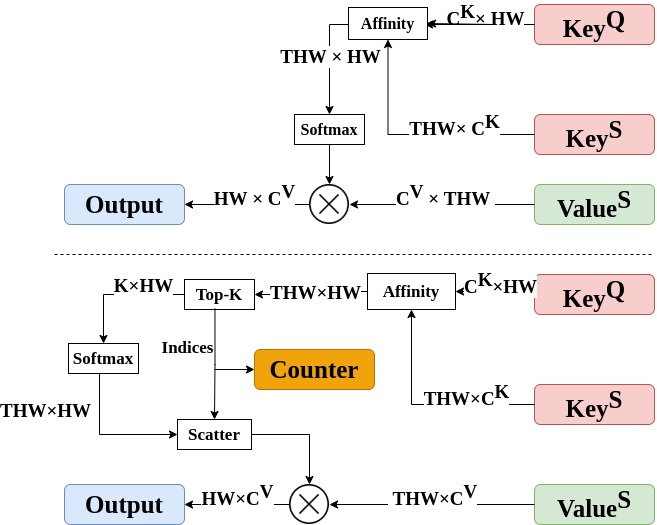}
  \caption{\textbf{Top:} Memory read block during training. H and W are the \textit{res4} layer dimensions which have a stride of 16 on the input. \textbf{Bottom:} Memory read block during inference. After finding the affinity between query and support keys, we find the k-closest features for each query feature. The counter module keeps track of the support features' appearance in top-k. }
  \label{fig:memory}
\end{figure}

\vspace{-5pt}
A memory read is a visual attention operation to reconstruct the support value with respect to the affinity between the support key and the query key. Figure \ref{fig:memory} shows an overview of our memory read module. After encoding, the key-value pairs $(K^S,V^S)$ of the support frames are concatenated to form a space-time key-value. A memory read operation starts by calculating the affinity between the support key $K^S \in \mathbb{R}^{THW \times C^K}$ and the query key $K^Q \in \mathbb{R}^{HW \times C^K}$. $T$ refers to number of frames in the support set. The affinity $d_{ij}$ between a support feature  $K^{S}_{i}$ and a query feature $K^{Q}_{i}$ is based on the negative squared distance and is given by,
\begin{align}
    d_{ij} &= \frac{dist(K^{S}_i,K^{Q}_j)}{\sqrt{C^k}}
    \label{eq:affinity}
\end{align}
where $dist(.,.):\mathbb{R}^{C^k} \times \mathbb{R}^{C^k} \longrightarrow  \mathbb{R}$ is the negative squared distance i.e. $- \Vert K^{S}_i,K^{Q}_j \Vert^{2}_{2}$. Similar to \cite{oh2019video,cheng2021rethinking}, the affinity $d_{ij}$ is divided by $\sqrt{C^k}$. %; not shown for brevity. 

The affinity matrix $D \in \mathbb{R}^{THW \times HW}$ is then normalized along the dimension of the query features using Softmax and is used to calculate $D \in \mathbb{R}^{THW \times C^V}$ as the weighted summation of the support value.

\subsection{Decoder}
\vspace{-5pt}
The decoder extends the work in STM \cite{oh2019video}. It contains two consecutive refinement modules \cite{oh2018fast} which upsample the features' spatial size by four and reduce the number of channels from 1024 to 1. The output of the decoder is then upsampled by four to match the size of the input. At each step, we concatenate the features of the query key encoder with the input through skip-connections. In the first stage, we reduce the number of channels of skip-connections to match the number of channels of memory read output using a $3 \times 3$ convolution layer. In the case of multiple objects, soft aggregation \cite{oh2019video} is used to reach the final mask.

\subsection{Memory bank}
\vspace{-5pt}
During the inference, we use a memory bank to store the support frames' information. The size of the memory bank can grow with the length of the video. Storing all previous frames imposes significant memory requirements and slows down performance during inference.

\begin{figure}[!ht]
    \centering
  \includegraphics[width=0.9\linewidth]{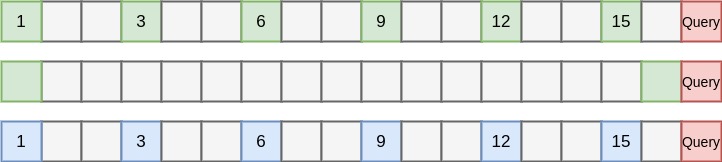}
  \caption{Various memory management strategies. \textbf{Top:} Storing every-k frames into the memory. The size of the memory bank grows linearly with the length of the video. \textbf{Middle:} Storing first-and-latest frames into the memory. Although this method uses a fixed-size memory bank, it disregards all intermediate frames. \textbf{Bottom:} We store every-\textit{k} frames into the memory. However, we remove obsolete features from the memory to maintain a fixed-size memory bank.}
  \label{fig:strategies}
\end{figure}

To address this limitation, STM \cite{oh2019video} suggested storing only the first-and-latest frames to maintain a fixed-size memory bank. However, experiments show that using intermediate frames improves the performance of the model. To benefit from intermediate frames, STM \cite{oh2019video} stores every k frame ($k=5$ in their experiments) into the memory bank. On an NVIDIA 1070 with 8G of memory, STCN \cite{cheng2021rethinking} can handle only ~100 frames in the memory bank, which is equivalent to a 15-second \textit{story} on Instagram. In order to handle long videos, streams with arbitrary length, or to use embedded devices with limited resources, the first-and-latest strategy offers the best solution. Figure \ref{fig:strategies} shows different memory management strategies.

% MiVOS \cite{cheng2021mivos} shows that it is not necessary to store all features. Only a fraction of the support features is involved in the segmentation process at each time.  

MiVOS \cite{cheng2021mivos} shows that after applying softmax to the affinity matrix between support and query key, the weights for most of the support features become small. Feature with small weights does not meaningfully contribute to the segmentation process. This phenomenon amplifies as the number of support features increases. MiVOS \cite{cheng2021mivos} shows that disregarding non-contributing features of the support set leads to more stable segmentation through time.

To the best of our knowledge, only the Adaptive Feature Bank (AFB) presented in \cite{liang2020video} attempts to address this problem. However, this method uses a user-defined threshold which is data-dependent and requires adjustment for different videos. More importantly, it supports only dot product or cosine similarity as a distance metric which, as shown in \cite{cheng2021rethinking}, are outperformed by the negative Euclidean distance.

We propose storing only the top-k features by updating the memory bank and removing obsolete features to overcome these limitations. The advantages are threefold: it eliminates the threshold requirement, results in a constant fixed-sized memory bank, and is agnostic to the distance metric used.

\subsubsection{Storing the top-\textit{k}}
As the number of features in the memory bank grows, only a fraction of those features will continue to be relevant and hence will have non-zero values after the softmax operation \cite{cheng2021mivos}. Additionally, frames that are temporally closer to the query frame are more likely to be visually similar. From the many cache management algorithms, the least-frequently-used (LFU) policy defined as $LFU = \frac{index}{age}$ is the most suitable for this task. Calculating the LFU score requires the index and the age. The index is calculated as the number of times that the feature has been referenced, i.e. the number of times it appeared in the top-\textit{k} features matching a query feature. The age is the time and is calculated in terms of the number of frames that the feature has been in the memory. 

%We calculate the index using a softmax normalized affinity matrix that promotes robust discriminative features and removes others. This is crucial to segment boundaries correctly and does not impose any restrictions on the distance metrics. 

%  Also, we have experimented with least-recently used (LRU) and LFU with dynamic aging. LRU is not suitable since it favors recent frames feature since they are close to the query frame. Also LFU with dynamic aging does not bring any improvement.
Unlike \cite{liang2020memory}, no updates are performed when adding a new frame to the memory. During removal, we remove enough features to make accommodate new features.

\begin{table}[!ht]
\centering
\caption{Quantitative results on DAVIS 2016 and DAVIS 2017 validation sets. Evaluation is performed with J(Intersection-over-Union) and F(boundary accuracy) metrics. Mean is the average of individual objects J and F. Recall measures objects scoring higher than a threshold ($\tau = 0.5$ following the official benchmarks \cite{perazzi2016benchmark}) in a sequence. Decay indicates the performance drop between the first-and-latest quartile of a sequence.}
\resizebox{0.45\textwidth}{!}{%
\begin{tabular}{cccccccc}
\hline
\multicolumn{1}{c|}{Methods}          & J-Mean & J-Recall & J-decay & F-Mean & F-Recall & \multicolumn{1}{c|}{F-decay} & J\&F-Mean \\ \hline
                                      & \multicolumn{7}{c}{DAVIS 2016 validation set}                                              \\ \hline
\multicolumn{1}{c|}{Every 5}          & 90.4   & 98.1     & 4.1     & 93.0   & 97.1     & \multicolumn{1}{c|}{4.3}     & 91.7      \\
\multicolumn{1}{c|}{first-and-latest} & 88.0   & 95.3     & 8.3     & 91.0   & 94.4     & \multicolumn{1}{c|}{8.4}     & 89.5      \\
\multicolumn{1}{c|}{Ours}             & 89.5   & 97.6     & 6.1     & 92.3   & 96.8     & \multicolumn{1}{c|}{5.8}     & 90.9      \\ \hline
                                      & \multicolumn{7}{c}{DAVIS 2017 validation set}                                              \\ \hline
\multicolumn{1}{c|}{Every 5}          & 82.0   & 91.3     & 6.2     & 88.6   & 94.6     & \multicolumn{1}{c|}{8.6}     & 85.3      \\
\multicolumn{1}{c|}{first-and-latest} & 79.5   & 90.4     & 10.8    & 85.4   & 92.9     & \multicolumn{1}{c|}{14.0}    & 82.5      \\
\multicolumn{1}{c|}{Ours}             & 81.3   & 90.4     & 8.1     & 87.4   & 92.9     & \multicolumn{1}{c|}{10.7}    & 84.4      \\ \hline
\end{tabular}%
 }
\label{tab:davis}
\end{table}

% TODO: can add davis2016 results (davis 2017 is better indicator anyway) Done!
\begin{table}[!ht]
\centering
\caption{Quantitative results on Youtube-VOS validation set. Final \textit{J\&F} score is unweighted average between seen and unseen metrics.}
% \resizebox{\columnwidth}{!}{%
\begin{tabular}{c|cccc|c}
\hline
Methods        & J-seen & J-unseen & F-seen & F-unseen & J\&F \\ \hline
every 5        & 82.6   & 79.3     & 86.9   & 87.6     & 84.1 \\
First-and-latest & 79.9   & 75.1     & 83.9   & 83.2     & 80.5 \\
Ours           & 80.3   & 76.1       & 84.3   & 83.8     & 81.1 \\ \hline
\end{tabular}
% }
\label{tab:ytbvos}
\end{table}

\vspace{-10pt}
\section{Implementation Details}
\vspace{-5pt}
We followed the training procedure in \cite{cheng2021rethinking} as training is not the main focus of our work. As suggested in \cite{cheng2021rethinking, oh2019video,cheng2021mivos}, we used two-stage training. First, we train the model on static images \cite{zeng2019towardsHRSOD, cheng2020cascadepsp, FSS1000, wang2017learning, shi2015hierarchicalECSSD} with augmented deformations for 300,000 iterations with batch size of 16. 

For the next stage, we use Youtube-VOS \cite{xu2018youtubeVOS}, and DAVIS \cite{perazzi2016benchmark,pont20172017} to train the network for 150,000 iterations with a batch size of 8. In this stage, at each iteration, we pick three temporally ordered frames. Following \cite{oh2019video,yang2020collaborative}, we use the first frame to segment the second frame and use second frame predictions and the first frame to segment the third frame.

We used Adam \cite{kingma2014adam} as optimizer, bootstrap crossentropy \cite{cheng2021mivos} as loss function and used 4 \textit{P100} GPUs to train the model which took 5 days to complete. We used Pytorch \cite{paszke2019pytorch} as a deep learning framework. 

For inference, we used a \textit{GTX1070} GPU and re-time STCN \cite{cheng2021rethinking} to ensure fair comparisons in our experiments. Given that support keys are object-agnostic, we initialize a single index and age counter at the beginning of the sequence, which is shared by all objects. In Youtube-VOS dataset \cite{xu2018youtubeVOS}, objects not always appear in the first frame. Using LFU, our method can successfully handle new objects appearing in the middle of the sequence.

% \begin{comment}
% \begin{table*}[!htb]
% \centering
% % \resizebox{\columnwidth}{!}{%
% \begin{tabular}{c|cccccc|c}
% \hline
% Methods          & J-Mean & J-Recall & J-Decay & F-Mean & F-Recall & F-decay & J\&F-Mean \\ \hline
% Every 5          & 90.4   & 98.1     & 4.1     & 93.0   & 97.1     & 4.3     & 91.7      \\
% first-and-latest & 88.0   & 95.3     & 8.3     & 91.0   & 94.4     & 8.4     & 89.5      \\
% Ours             & 89.5   & 97.6     & 6.1     & 92.3   & 96.8     & 5.8     & 90.9      \\ \hline
% \end{tabular}
% %  }
% \caption{Quantitative results on DAVIS 2016 validation.}
% \label{tab:davis16}
% \end{table*}

% \begin{table*}[!ht]
% \centering
% % \resizebox{\columnwidth}{!}{%
% \begin{tabular}{c|cccccc|c}
% \hline
% Methods        & J-Mean & J-Recall & J-Decay & F-Mean & F-Recall & F-Decay & J\&F-Mean \\ \hline
% Every 5        & 82.0   & 91.3     & 6.2     & 88.6   & 94.6     & 8.6     & 85.3      \\
% first-and-latest & 79.5   & 90.4     & 10.8    & 85.4   & 92.9     & 14.0    & 82.5      \\
% Ours           & 81.3   & 90.4     & 8.1     & 87.4   & 92.9     & 10.7    & 84.4      \\ \hline
% \end{tabular}
% % }
% \caption{Quantitative results on DAVIS 2017 validation.}
% \label{tab:davis17}
% \end{table*}
% \end{comment}

\vspace{-10pt}
\section{Experiments}
\vspace{-5pt}
We evaluate our approach on DAVIS 2017 \cite{pont20172017} validation set, and Youtube-VOS \cite{xu2018youtubeVOS}, two large-scale benchmarks with multiple objects in videos. In DAVIS 2017, all target objects are present in the first frame of the sequence. However, they can be occluded at the beginning or disappear and reappear in the middle of the sequence. In Youtube-VOS, some of the target objects first appear in the middle of the video, and the objective is to start tracking that object from that point onwards. 

For evaluation, we use \textit{J\&F} from the DAVIS benchmark, which is an average between the region accuracy \textit{J} and the boundary accuracy \textit{F}. For each object, we calculate \textit{J\&F} score in each frame separately, and the object score is the mean of its score in different frames. The overall score is an average of each object score. This method prevents big objects or objects with more extended visibility from skewing the results. 

We evaluate our approach with every-\textit{k} and first-and-latest methods. In all experiments, we ensure a fair comparison with the first-and-latest approach by setting the size of the memory bank to two frames worth of features. We compare our method with first-and-latest memory utilization since it is the sole available method that can handle videos of arbitrary length. Additionally, we compare our approach with every-\textit{k} since it is the best performing approach for memory write. Results show that our approach outperforms first-and-latest and has comparable results with the every-\textit{k} method. We also investigate the memory utilization and inference speed of different approaches. Our method has better inference speed with minimal memory utilization. 

\noindent
\textbf{Results on DAVIS 2016.} DAVIS 2016 \cite{perazzi2016benchmark} validation set contains 20 videos with dense masks for single objects. A quantitative comparison of our approach with other methods is shown in Table \ref{tab:davis}. Our network outperforms the first-and-latest method by \textbf{1.4 \%}. To evaluate the network's ability to handle multi-object scenarios closer to real-world applications, we next present the results on DAVIS 2017.

\noindent
\textbf{Results on DAVIS 2017.} DAVIS 2017 is a multi-object extension of DAVIS 2016 dataset. It contains 120 videos that are 30 times smaller than the videos in the Youtube-VOS dataset. The validation set has 59 objects in 30 videos. The results on DAVIS 2017 are shown in Table \ref{tab:davis}. Our method outperforms the first-and-latest method by a significant margin and has comparable results with every-\textit{5} method while using a smaller memory bank.

\begin{figure*}[!ht]
  \centering
  \includegraphics[width=0.85\textwidth]{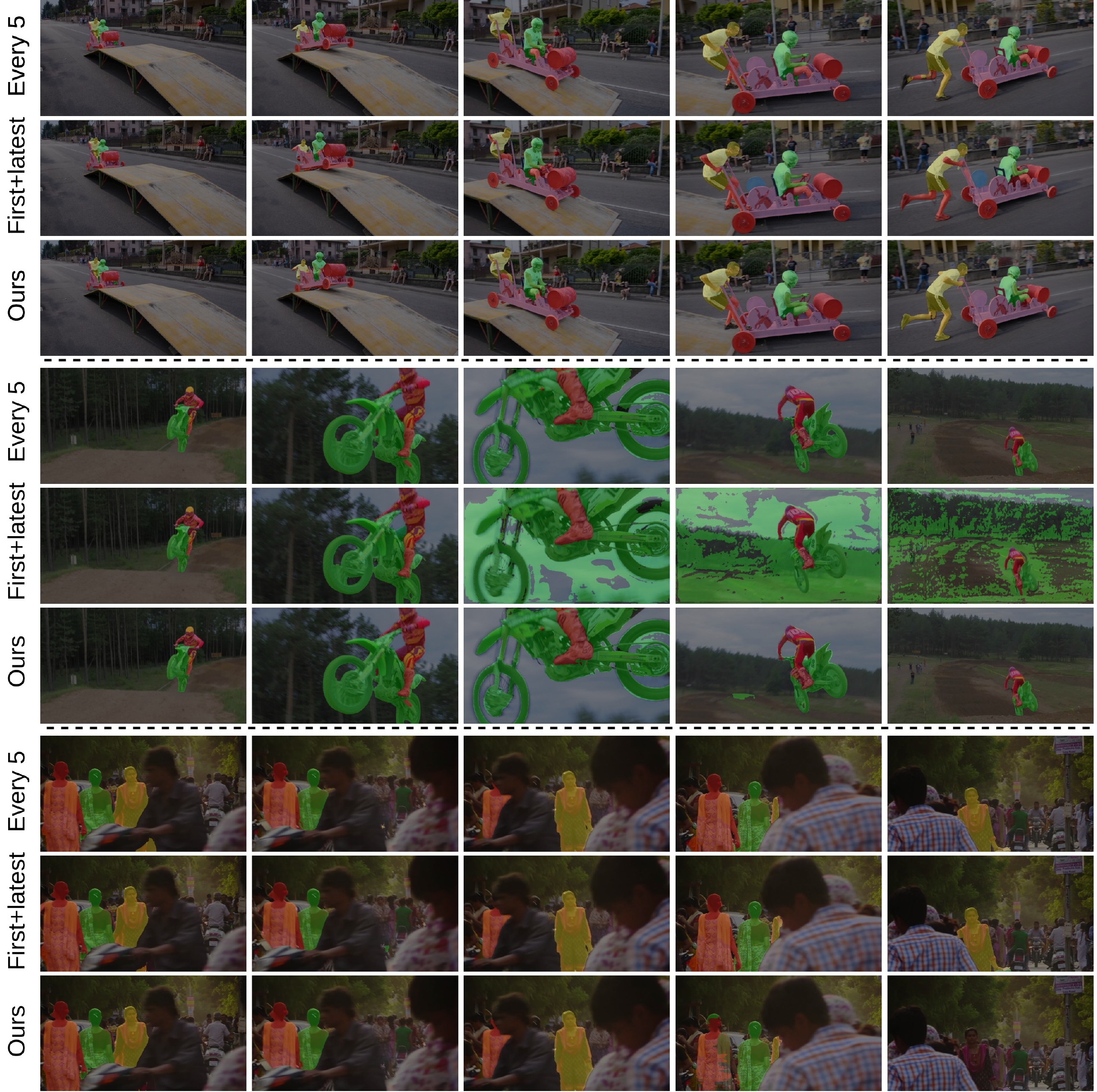}
  \caption{Qualitative results for DAVIS 2017. Frames are sampled from DAVIS 2017 \cite{pont20172017} validation set. In each row, frames are temporally ordered from left to right. Frames are sampled from challenging situations and transitions. \textbf{Top:} Our method can successfully recover from the drifting. \textbf{Middle:} First-and-latest approach collapses as a result of fast object deformation. \textbf{Bottom:} Our method fails to re-identify object that has been completely occluded for a few frames. Features that belonged to this object got removed after being unused for a few frames. Since object form has not been changed through time, the first-and-latest method can successfully segment using first frame information.}
  \label{fig:davis17}
\end{figure*}

\begin{figure*}[!ht]
    \centering
  \includegraphics[width=0.85\textwidth]{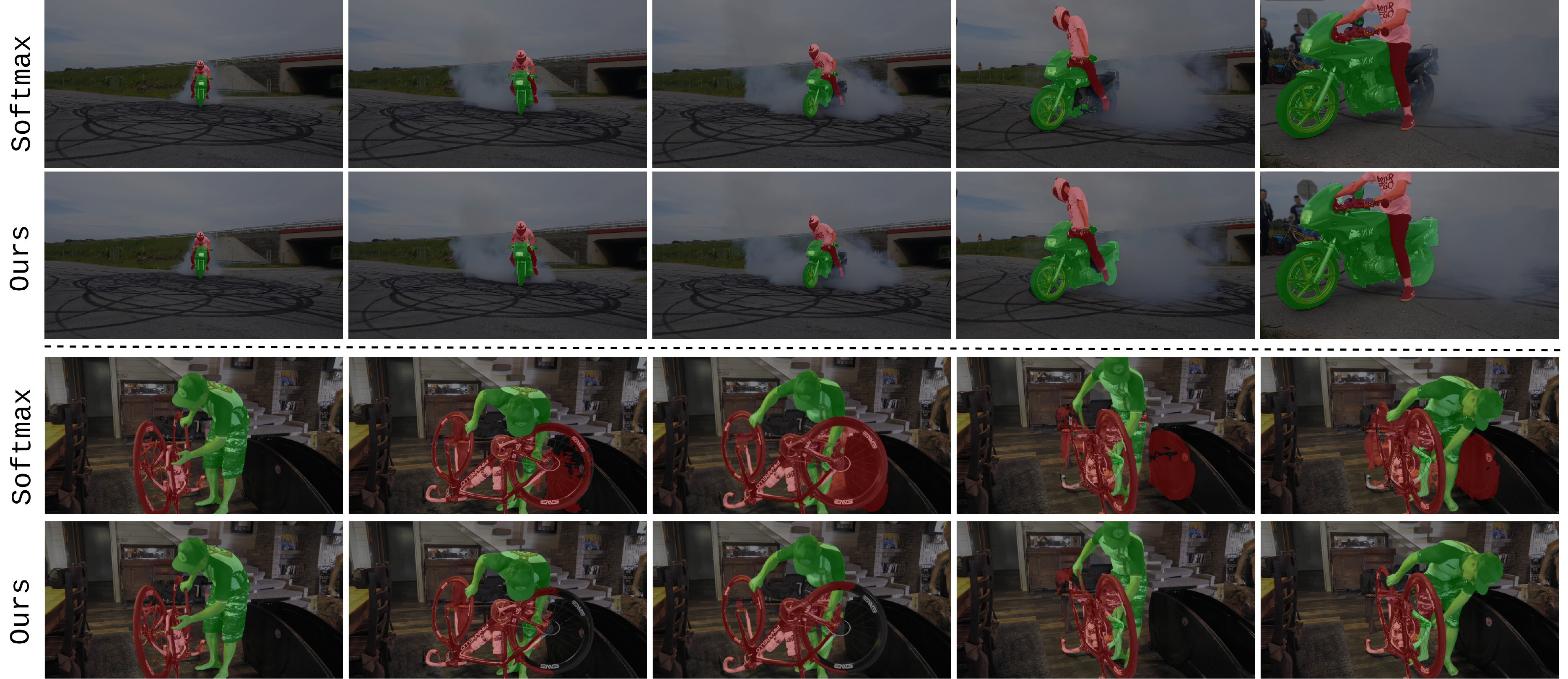}
  \caption{Qualitative results for top-k and softmax weights as an index. Frames are sampled from DAVIS 2017 \cite{pont20172017} validation set. In each row, frames are temporally ordered from left to right. Frames are sampled from challenging situations and transitions. \textbf{Top:} top two rows shows top-k effectiveness to handle deformation of the object. \textbf{Bottom:} In the two bottom rows, we can see that using softmax weights, the network is unable to adapt to the object's new appearance as it changes toward the end of the sequence.}
  \label{fig:softmax}
\end{figure*}

\begin{table}[ht]
\centering
\caption{Comparison of Inference speed and memory utilization of different memory management strategies on DAVIS 2017 validation set.}
% \resizebox{\textwidth}{!}{%
\begin{tabular}{c|ccc} \hline
                   & \multicolumn{1}{c}{Every 5} & First-and-latest & Ours  \\ \hline
Frame per Second                & 6.3                         & 8.4            & 11.4 \\ \hline
Memory utilization & 7                           & 2              & 2   \\ \hline
\end{tabular}%
% }
\label{tab:time}
\end{table}

\noindent
\textbf{Results on Youtube-VOS.} Youtube-VOS is the latest large-scale dataset for VOS. The training set has 3471 videos with 65 different object categories. The validation set has 507 videos with 26 unseen object categories and the object categories of the train set. The availability of unseen categories makes Youtube-VOS suitable to measure the generalization ability of the various methods in question. The results for Youtube-VOS are shown in Table \ref{tab:ytbvos}. Our method outperforms the first-and-latest method and underperforms when compared with every-\textit{k} approach. Since the average length of videos for Youtube-VOS is longer than DAVIS2017, using the same every-\textit{k} setup, the Youtube-VOS causes increased memory use; it results in storing a more significant number of frames in the memory bank, which consequently makes it harder to compete with using only two frames worth of features.

\noindent
\textbf{Inference speed.} The size of the memory bank directly affects the amount of computation in the memory read block. By limiting the size of the memory bank, the proposed method can increase the inference speed by 80 \%. A comparison of inference speed and memory utilization is shown in Table \ref{tab:time} (The inference speed shows the number of frames processed in a second in a multi-object video. Memory utilization shows the number of frames stored in the memory bank. For the every-5 method, we used the average number of frames in the memory. In our method, we set the size of the memory bank to be equal to 2 frames worth of features.) The first-and-latest approach stores the latest frame into the memory at each step which slows down the speed. On the other hand, our method can be described as an extension to the every-\textit{k} meaning that we only perform a memory write operation every few frames. Having fewer memory write operations leads to an increase in inference speed by 35\%. To calculate the inference speed, we measure the total processing time on the whole DAVIS 2017 \cite{pont20172017} validation set and divide it by the total number of frames.

\begin{table}[!ht]
\centering
\caption{Ablation on top-k storing versus softmax weights storing}
\resizebox{0.45\textwidth}{!}{%
\begin{tabular}{c|ccc|ccc}
\hline
\multirow{2}{*}{Methods} & \multicolumn{3}{c|}{DAVIS 2017} & \multicolumn{3}{c}{Youtube-VOS} \\ \cline{2-7} 
                         & J-Mean   & F-Mean  & J\&F-Mean  & J-Mean   & F-Mean  & J\&F-Mean  \\ \hline
Softmax storing          & 79.9     & 82.2    & 83.0       & 76.2     & 81.8    & 79.0       \\
Top-k storing            & 81.3     & 87.4    & 84.4       & 78.2     & 84.0    & 81.1       \\ \hline
\end{tabular}%
}

\label{tab:ablation}
\end{table}

\noindent
\textbf{Qualitative results} are shown in Figure \ref{fig:davis17}. The first six rows show how different approaches handle multiple objects exhibiting deformation and significant displacement due to motion. As we do not apply any spatial constraint to the segmentation, the network can handle significant displacements successfully. However, in the case of deformation, the object's appearance in the initial frames can quickly become distant -in metric space- from that in the current frame. In this case, the network relies heavily on the mask propagated from the previous frame. This makes the network prone to accumulating errors over time. On the other hand, our strategy can handle deformation and recover from erroneous previous predictions by removing obsolete features from memory. 

The bottom three rows in Figure \ref{fig:davis17} show the case of complete occlusion when tracking multiple objects. Given the object-agnostic nature of our baseline, we are using the same index counter for all of the objects. During an object's occlusion, features become inactive. As a result, they have a higher probability of being discarded from memory, and in the case they reappear later in the video sequence, the network may not track the object again.

\noindent
\textbf{Discussion.} We analyzed the effectiveness of using top-k as an index. To do so, we remove the top-k storing from the memory bank and instead use softmax weights as the counter index. A comparison between the two approaches is shown in Table \ref{tab:ablation}.

Even among the top-k features, softmax normalization is highly imbalanced toward the closest features, which score a high probability. This imbalance diminishes the network's ability to discard features that used to be deterministic but lost their effectiveness. Figure \ref{fig:softmax} shows a qualitative comparison between top-k and softmax weights used as the index. As seen, our method can handle the deformation and complex motions of objects.

\begin{figure}[!ht]
 \centering
  \includegraphics[width=0.3\textwidth]{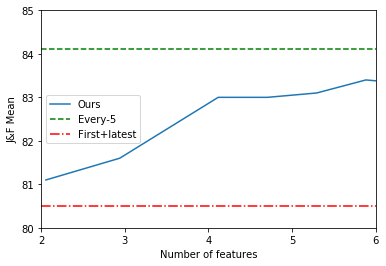}
  \caption{Effect of memory bank capacity on \textit{J\&F} metric. Every-5 and first-and-latest approach performance are shown for a clearer comparison. The size of the memory bank is specified in terms of the number of frames.}
  \label{fig:size}
\end{figure}

\noindent
\textbf{Memory bank size.} We further investigate the effect of memory bank capacity on model performance. For this purpose, we gradually increase the size of the memory bank and evaluate the performance of the network on Youtube-VOS \cite{xu2018youtubeVOS} validation set. We use Youtube-VOS \cite{xu2018youtubeVOS} since it has larger validation set with longer videos. Results are visualized in Figure \ref{fig:size}. The results show that increasing the memory bank's capacity leads to a better \textit{J\&F} score. More importantly, only by doubling memory bank capacity from 2 to 4 frames worth of features, the network's  \textit{J\&F} measure increases by 2\%. The results show that using a memory bank with a capacity of four frames worth of features is the best trade-off between inference speed and accuracy for the Youtube-VOS dataset.

\vspace{-10pt}
\section{Conclusion}
\vspace{-5pt}
We presented a memory management strategy for semi-supervised video object segmentation. We employed a least-frequently-used(LFU) policy using the top-\textit{k} index. Extensive experimentation on DAVIS 2016, DAVIS 2017, and Youtube-VOS demonstrates that our method outperforms fist-and-latest strategies with a fixed-sized memory bank and achieves comparable results with every-\textit{k} strategies with an increasing-sized memory bank. Unlike state-of-the-art every-\textit{k} methods, ours handles videos of arbitrary length with no additional overhead, which is crucial for real-world applications.  Furthermore, our method facilitates video object segmentation of arbitrary-lengthed video streams under limited computational resources.

\vspace{-10pt}
\section*{Acknowledgment}
\vspace{-5pt}
This research is supported in part by the Natural Sciences and Engineering Research Council of Canada Grants DGN01670 (Discovery Grant) and DND-N01885 (Collaborative Research and Development with the Department of National Defence Grant).



%\bibliographystyle{IEEEtran}


% \end{thebibliography}
\vspace{-5pt}
{\tiny
\bibliographystyle{IEEEtranBST/IEEEtran}
\bibliography{IEEEtranBST/IEEEabrv,IEEEtranBST/main}
}

\clearpage
\section*{SUPPLEMENTARY MATERIAL}

% \usepackage{graphicx}
\begin{table*}[t]
\centering
\caption{Qualitative results on DAVIS 2016 and DAVIS 2017 validation sets. For evaluation, we followed the original STCN \cite{cheng2021rethinking} strategy to ensure a fair comparison.}
% \resizebox{\textwidth}{!}{%
\begin{tabular}{cccccccc}
\hline
\multicolumn{1}{c|}{Methods}             & J-Mean & J-Recall & J-decay & F-Mean & F-Recall & \multicolumn{1}{c|}{F-decay} & J\&F-Mean \\ \hline
                                         & \multicolumn{7}{c}{DAVIS 2016 validation set}                                              \\ \hline
\multicolumn{1}{c|}{STCN + dropout}      & 90.0   & 98.1     & 4.4     & 92.8   & 97.2     & \multicolumn{1}{c|}{4.9}     & 91.4      \\
\multicolumn{1}{c|}{STCN + reallocation} & 89.7   & 97.3     & 4.4     & 91.2   & 96.4     & \multicolumn{1}{c|}{5.5}     & 90.5      \\
\multicolumn{1}{c|}{STCN}                & 90.4   & 98.1     & 4.1     & 93.0   & 97.1     & \multicolumn{1}{c|}{4.3}     & \textbf{91.7}      \\ \hline
                                         & \multicolumn{7}{c}{DAVIS 2017 validation set}                                              \\ \hline
\multicolumn{1}{c|}{STCN + dropout}      & 79.1   & 88.1     & 9.4     & 85.5   & 92.3     & \multicolumn{1}{c|}{10.5}    & 81.6      \\
\multicolumn{1}{c|}{STCN + reallocation} & 78.5   & 86.7     & 9.5     & 84.4   & 91.4     & \multicolumn{1}{c|}{11.0}    & 81.4      \\
\multicolumn{1}{c|}{STCN}                & 82.0   & 91.3     & 6.2     & 88.6   & 94.6     & \multicolumn{1}{c|}{8.6}     & \textbf{85.3}      \\ \hline
\end{tabular}%
% }

\label{tab:davis_extra}
\end{table*}

\begin{table*}[ht]
\centering
\caption{Quantitative results on Youtube-VOS validation set.}
% \resizebox{\columnwidth}{!}{%
\begin{tabular}{c|cccc|c}
\hline
Methods        & J-seen & J-unseen & F-seen & F-unseen & J\&F \\ \hline
STCN + dropout & 81.0,& 76.9 & 84.9 & 84.0 & 81.6 \\
STCN + reallocation & 80.9 & 75.8 & 85.2 & 83.6 & 81.4 \\
every 5        & 82.6   & 79.3     & 86.9   & 87.6     & \textbf{84.1} \\ \hline
\end{tabular}
% }
\label{tab:ytbvos_extra}
\end{table*}

\section{Memory bank regularization}
During segmentation, we establish a pixel-wise connection between features of the support key and query key. As mentioned in \cite{cheng2021rethinking,cheng2021mivos}, the relation between support features and query features can be dominated by a small portion of discriminant features, which essentially limits other features information to be propagated. STCN \cite{cheng2021rethinking} addresses this issue by replacing the dot product with negative Euclidean distance as the affinity metric since the dot product is biased toward activated features with a more significant norm. This section applies two regularization methods to the affinity matrix and studies their effect on the performance. The intuition is to reduce network reliance on discriminant features and improve network robustness to occlusions or deformations. To address this issue, we investigate dropout and support feature reallocation in the memory block.  

Dropout was initially proposed to regularize the network during training and prevent it from over-fitting. Dropout randomly deactivates features during the training and makes the network perform the task using the remaining information. This operation increases the generalization ability and leads to more robust performance. After calculating the affinity matrix, we introduce a Dropout layer to randomly disable feature connection using binomial distribution $Pr(q)$ with $q=0.5$. Dropout shifts attention to remaining features and forces the network to perform segmentation based on remaining information. This simulates occlusion or illumination change where query and support connection is affected by abrupt changes. 

In addition to dropout, inspired by \cite{wang2020few}, we also investigate the effect of support feature reallocation on network performance. The main goal is to reallocate the network's attention to a broader range of support features. The affinity matrix $D$ shows pair-wise relation between query key $K^q$ and support key $K^s$ as is demonstrated in Eq. 1 in the manuscript. By taking the average along support key dimension, we create an attention map for support features.
\begin{align*}
    d_{ij} &= dist(k^{s}_i,k^{q}_j) \\
    A_{i} &= \sum_{j=1}^{HW} d_{ij}
\end{align*}
The attention value of each support feature reflects its importance during the segmentation. We reallocate the attention between support features to facilitate information propagation and include more support features in the segmentation process. To do so, we sort support feature attention in descending order and get sorting index $w$ for each feature $i$ in support. Having sorted the index, we reweight the connection belonging to feature $i$ in the affinity matrix using the following equation.
\begin{align*}
    \hat{d_{ij}} = d_{ij} \times \frac{w_j}{\sum_{k=1}^HW w_k}
\end{align*}
Multiplying normalized sorted index to corresponding pixels of affinity matrix increases the importance of the less discriminant features. Since most of the pixels belong to the background, we only perform this operation on the pixels of each object separately. For each object in the support set, we store the corresponding object mask and multiply interpolated mask to support key to filter out background features.

We followed the training procedure suggested by STCN \cite{cheng2021rethinking} to study the effect of these two regularizations on the network. We evaluated each method after complete training on Youtube-VOS \cite{xu2018youtubeVOS}, DAVIS 2017 \cite{pont20172017} and DAVIS 2016 \cite{perazzi2016benchmark}. Quantitative Results for DAVIS 2017 and 2016 are shown in table \ref{tab:davis_extra} and Quantitative results for Youtube-VOS are shown in table \ref{tab:ytbvos_extra}.

On DAVIS 2016 \cite{perazzi2016benchmark} single object benchmark, both regularization methods achieve close results to the baseline. However, neither of them can improve the performance of the method. On DAVIS 2017\cite{pont20172017} and Youtube-VOS \cite{xu2018youtubeVOS} multi-object datasets, the network performance degrades using these regularization methods.

We randomly pick two objects from the sequence for multi-object training to save computation and reduce complexity in each iteration. However, both dropout and feature reallocation are harmful to multi-object training and add unproductive complexity to the network.

\section{Training details}
We follow \cite{cheng2021rethinking,cheng2021mivos} strategies for data augmentation and training. They are mentioned here for completeness. However, we invite readers to look at the open-source code for more details.

\subsection{Pre-training}
At the pre-training stage, we used BIG\cite{cheng2020cascadepsp}, DUTS\cite{wang2017learning}, HRSOD\cite{zeng2019towardsHRSOD}, FSS1000\cite{FSS1000}, and ECSSD\cite{shi2015hierarchicalECSSD}. BIG and HRSOD repeated five times in training data as they provide higher quality annotations. At each iteration, we randomly pick one image and generate three augmented samples from the base image. Before creating augmented samples, we apply random scaling, random horizontal flip, random color jitter, and random greyscale on the base image using Pytorch augmentation tools. To create augmented samples, we apply affine transform and another color jitter on the augmented base image. Each sample shorter side is then resized to 384, and a random crop applied after that resulting in $384 \times 384$ outputs. In this stage, we only train the network with single object samples.

\subsection{Main-training}
At the main-training stage, we use Youtube-VOS \cite{xu2018youtubeVOS}, and DAVIS \cite{pont20172017,perazzi2016benchmark}. DAVIS is repeated five times as it contains better annotations. We used the 480p version of DAVIS for training and resized Youtube-VOS such that the shorter edge is equal to 480. At each iteration, we sample three temporally ordered frames from a video to form the training sequence. The maximum temporal distance of sampled frames starts from 5 and increases by 5 at $[10\%, 20\%, 30\%, 40\%$. For the last $10\%$ of iterations, the maximum temporal distance is changed back to 5 frames. 

For augmentation, we apply random horizontal flip, random resized crop ($crop\_size=384$), random color jitter and random grayscale on all images in the training sequence. The random seed is fixed for all images in the training sequence, and we apply the same transform on all images. Then, for each image, we perform color jitter and random affine transform. At each training sequence, we pick at most two objects to perform multi-object training. 

\subsection{Loss, Optimizer, and scheduling}
Adam \cite{kingma2014adam} optimizer is initialized with base learning rate of $10^-5$, momentum of $\beta_1=0.9,\beta_2=0.999$, L2 weight decay of $10^-7$, and step learning rate decay with decay ratio of 0.1. Decay is performed once in the middle of the pre-training stage and once in the middle of the main training. The learning rate schedule ends after each stage of training, and we initialize a new schedule for each stage.
Batch normalization layers in the key and value encoder are frozen with the Imagenet pre-trained network parameters to speed up the training.
For bootstrapped cross-entropy loss, we used $100\%$ of pixels for the first 20k iterations. After that, we only use top-$p\%$ pixels with the highest error (hard pixel-mining) to compute gradients. $p$ linearly decreases from 100 to 15 over 50k iteration and remains fixed after that.

\end{document}